\documentclass{article}
\usepackage{spconf,amsmath,graphicx,hyperref}
\usepackage{amssymb}
\usepackage{multirow}

\title{MAR: Efficient Large Language Models via Module-aware Architecture Refinement}
\name{
\parbox{\linewidth}{\centering Junhong Cai\textsuperscript{1}, Guiqin Wang\textsuperscript{2,3}, Kejie Zhao\textsuperscript{1},  Jianxiong Tang\textsuperscript{4}, Xiang Wang\textsuperscript{5}, \\
Luziwei Leng\textsuperscript{5}, Ran Cheng\textsuperscript{6}, Yuxin Ma\textsuperscript{*}\textsuperscript{1}, Qinghai Guo\textsuperscript{*}\textsuperscript{5}\thanks{\textsuperscript{*}The corresponding authors are Yuxin Ma (mayx@sustech.edu.cn) and Qinghai Guo (guoqinghai@huawei.com).
 }
 }
 }
\address{
    \textsuperscript{1}\textit{Department of CSE, Southern University of Science and Technology}, Shenzhen, China\\
    \textsuperscript{2}\textit{School of Computer Science and Technology, Xi’an Jiaotong University}, Xi’an, China\\
    \textsuperscript{3}\textit{National Engineering Laboratory for Big Data Analytics, Xi’an Jiaotong University}, Xi’an, China\\
    \textsuperscript{4}\textit{Department of Computer Science, City University of Hong Kong}, Hong Kong, China\\
    \textsuperscript{5}\textit{ACS Laboratory, Huawei Technologies Co., Ltd.}, Shenzhen, China\\
    \textsuperscript{6}\textit{Department of Data Science and Artificial Intelligence, Department of Computing,}\\ \textit{Hong Kong Polytechnic University}, Hong Kong, China}
\begin{document}
\ninept
\maketitle
\begin{abstract}
Large Language Models (LLMs) excel across diverse domains but suffer from high energy costs due to quadratic attention and dense Feed-Forward Network (FFN) operations.
To address these issues, we propose Module-aware Architecture Refinement (MAR), a two-stage framework that integrates State Space Models (SSMs) for linear-time sequence modeling and applies activation sparsification to reduce FFN costs.
In addition, to mitigate low information density and temporal mismatch in integrating Spiking Neural Networks (SNNs) with SSMs, we design the Adaptive Ternary Multi-step Neuron (ATMN) and the Spike-aware Bidirectional Distillation Strategy (SBDS).
Extensive experiments demonstrate that MAR effectively restores the performance of its dense counterpart under constrained resources while substantially reducing inference energy consumption. Furthermore, it outperforms efficient models of comparable or even larger scale, underscoring its potential for building efficient and practical LLMs.
\end{abstract}
\begin{keywords}
Efficient Large Language Models, Spiking neural networks, Knowledge distillation, Linear attention
\end{keywords}
\section{Introduction}
\label{sec:intro}

In recent years, Large Language Models (LLMs) \cite{touvron2023llama,yang2025qwen3,liu2024deepseek} have shown remarkable generalization and adaptability across diverse domains \cite{huang2023lawyer,liu2023goat}. However, their massive parameter scales and high computational costs hinder both development and deployment.

\begin{figure}[t]
    \centering
    \includegraphics[width=0.9\columnwidth]{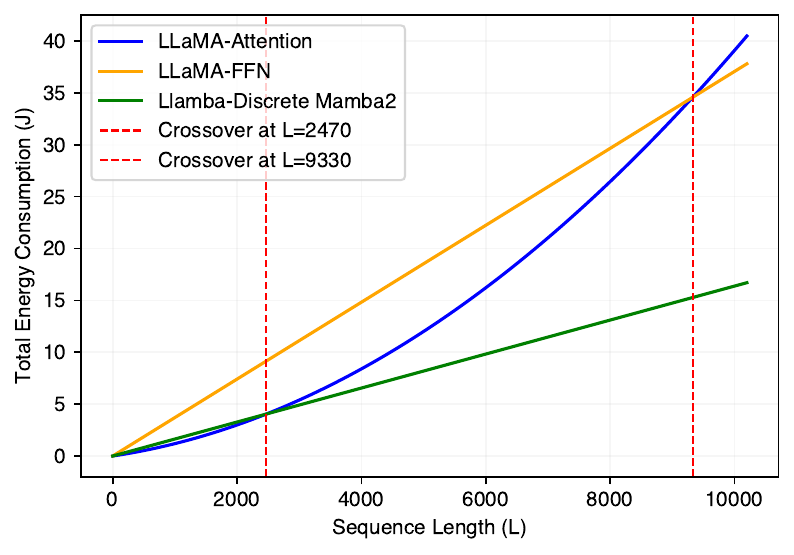}
    \vspace{-10pt}
    \caption{Energy consumption of LLaMA-3.2 Attention/FFN versus Llamba’s \cite{bick2025llamba} Discrete Mamba-2. Crossovers occur at sequence lengths 2470 and 9330.}
    \vspace{-5pt}
    \label{fig:MLP-berden}
\end{figure}

To mitigate these issues, research has pursued two main directions: (1) model compression techniques, such as quantization and knowledge distillation, to reduce energy consumption without altering architectures \cite{dettmers2022gpt3,xiao2023smoothquant,wang2025abkd}; and (2) architectural innovations to lower sequence modeling complexity, including FlashAttention variants \cite{dao2022flashattention,wang2020linformer} and State Space Models (SSMs) \cite{gu2021efficiently,gu2023mamba}. Yet, most efforts target the quadratic cost of attention \cite{vaswani2017attention}, while in non-extremely long sequences, Feed-Forward Networks (FFNs) often dominate energy consumption (see Figure \ref{fig:MLP-berden}). Thus, optimizing FFNs is equally critical for efficiency.

\begin{figure}
    \centering
    \includegraphics[width=1.0\linewidth]{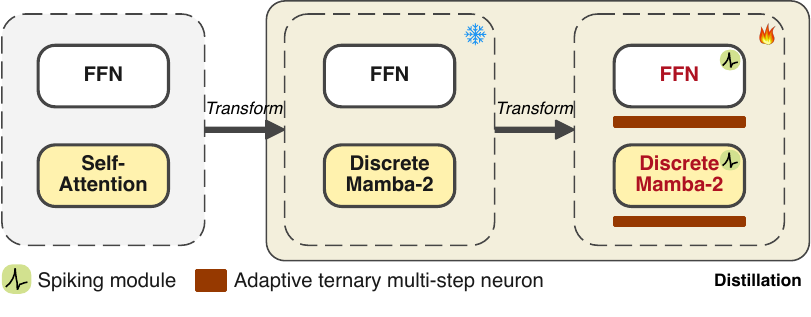}
    \vspace{-15pt}
    \caption{Overview of the MAR framework. This figure illustrates how a dense attention-based model is transformed into a sparse and efficient linear sequence model through a two-stage optimization process.}
    \vspace{-5pt}
    \label{fig:overview}
\end{figure}

Within architectural optimization, bio-inspired Spiking Neural Networks (SNNs) \cite{maass1997networks,tavanaei2019deep} have emerged as a promising solution to alleviate the computational burden of linear layers and reduce FFN energy consumption. Prior studies have shown that incorporating SNNs into Transformer architectures to sparsify activations can achieve encouraging results \cite{zhou2022spikformer}. However, extending this approach to SSMs remains non-trivial, primarily due to two challenges: (1) a temporal mismatch between the continuous dynamics of SSMs and the discrete nature of spike events, and (2) reduced information density, as the number of activations per timestep is notably lower than in conventional ANNs \cite{zhong2024spike}.

To overcome these limitations, we propose Module-aware Architecture Refinement (MAR), a two-stage framework that jointly linearizes attention mechanisms and reduces FFN computational cost. We further introduce the Spike-aware Bidirectional Distillation Strategy (SBDS) to preserve performance under resource constraints. The main contributions are as follows:

Firstly, to jointly optimize quadratic attention and FFNs, we propose the two-stage MAR. In the first stage, SSMs are introduced to achieve linear-time sequence modeling. In the second stage, spiking neurons are employed to sparsify activations, thereby substantially reducing the computational cost of linear layers.

Secondly, to address temporal mismatch and low information density, we propose Adaptive Ternary Multi-step Neuron (ATMN) and SBDS. ATMN increases information capacity to mitigate sparsity-induced loss, while SBDS uses reverse-KL compensation with Pre-Norm alignment to mitigate temporal mismatch.

Thirdly, extensive experimental results validate the effectiveness of our approach. Our linear spiking model successfully restores the performance of teacher model on multiple commonsense reasoning and question answering benchmarks while significantly reducing inference energy consumption. At the same time, it surpasses larger spiking LLM as well as other efficient models of comparable scale.

\section{Method}
\subsection{Module-aware Architecture Refinement}
To address the computational bottlenecks of attention mechanisms and FFNs, we propose Module-aware Architecture Refinement (MAR), a two-stage optimization framework that restructures LLMs through a module-aware strategy to improve the efficiency of both sequence modeling and FFN computation. An overview of the framework is shown in Figure \ref{fig:overview}.

In the first stage, we replace the attention mechanism with an SSM to achieve linear-time sequence processing. Given the maturity of prior work in this direction, we directly adopt Llamba \cite{bick2025llamba} as the baseline in the second stage. Llamba is an attention-free variant of LLaMA-3.2, built upon discrete multi-head Mamba-2 modules.

\begin{figure}[ht]
    \centering
    \includegraphics[width=0.9\columnwidth]{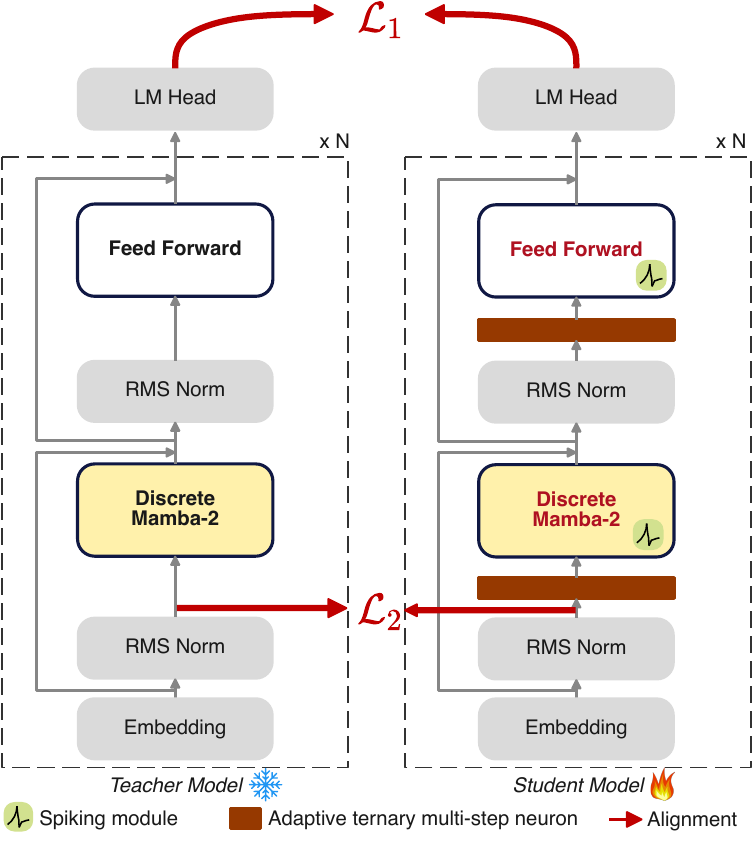}
    \vspace{-10pt}
    \caption{Overview of the SBDS: the student model (right) learns from the teacher model (left) through feature-level and logit-level losses.}
    \vspace{-5pt}
    \label{fig:distill}
\end{figure}

\begin{figure}
    \centering
    \includegraphics[width=1.0\columnwidth]{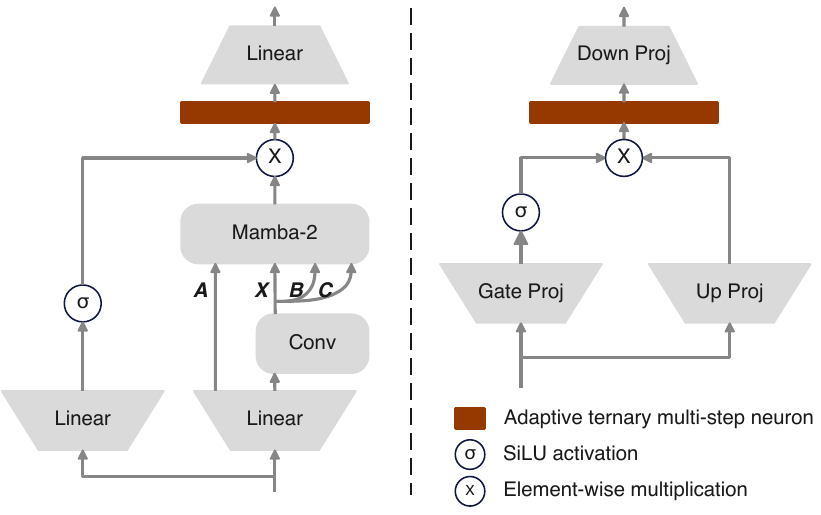}
    \vspace{-10pt}
    \caption{Illustration of spiking integration in the Discrete Mamba-2 (left) and FFN (right). The left and right sides correspond to the Discrete Mamba-2 and the Feed Forward Network in the student model shown in Figure \ref{fig:distill}, respectively.}
    \vspace{-5pt}
    \label{fig:spikingModule}
\end{figure}

In the second stage, MAR targets the high cost of FFNs and other fully connected layers. Spiking neurons (ATMNs) are inserted before fully connected units, sparsifying activations by replacing multiply–accumulate (MAC) with accumulate (AC) operations. In Llamba’s decoder, spiking neurons are placed at four positions per layer, before the input and output projections of both Mamba-2 and FFN modules (Figures \ref{fig:distill}, \ref{fig:spikingModule}).

\subsection{Adaptive Ternary Multi-step Neuron}
Traditional spiking neurons typically encode information using binary spike sequences with values in \{0, 1\}. However, this representation discards negative signals, thereby limiting the representational capacity of individual neurons and reducing information density. To overcome this limitation while preserving sparsity, we draw inspiration from the design of \cite{guo2024ternary} and propose the Adaptive Ternary Multi-step Neuron (ATMN), which enhances the information-carrying capacity of spiking neurons. Its working mechanism can be described by the following iterative equations:
\vspace{-5pt}
\begin{align}\label{atmn_eq1}
    h_t = I_t\cdot \delta_{t,0} + \frac{1}{\tau}\cdot u_{t-1}, 
\end{align}
\vspace{-5pt}
\begin{equation}\label{atmn_eq2}
    s_t=
    \begin{cases}
    1\quad \text{if }\ h_t \ge V_{adaptive}\\
    -1 \quad \text{if }\ h_t \le -V_{adaptive}\\
    0 \quad \text{otherwise},
\end{cases}
\end{equation}
\vspace{-5pt}
\begin{equation}\label{atmn_eq3}
    u_t = h_t- s_t\cdot V_{adaptive},
\end{equation}
\begin{equation}\label{atmn_eq4}
    V_{adaptive} = e^a,
\end{equation}
where $I_t$ denotes the input current at time step $t$, $\tau$ is the membrane time constant, and $h_t$, $u_t$ represent the membrane potentials before and after spike reset, respectively. $s_t$ indicates the spike output, and $V_{adaptive}$ is a neuron-wise adaptive threshold constrained to be non-negative via an exponential function with trainable parameter $a$. The Kronecker delta function $\delta_{t,0}$ equals 1 only at $t = 0$, allowing external input injection at the initial step.

Unlike conventional binary spiking neurons, ATMN emits a negative spike when the membrane potential falls below $-V_{adaptive}$, thereby mimicking biological inhibitory signals. This bidirectional spiking mechanism effectively mitigates the low information density issue caused by sparse activations and enhances representational capacity. In addition, we inject external input only at the initial step, while subsequent updates rely solely on the residual membrane potential, simplifying the temporal update process. During reset, the membrane potential moves toward the resting state if a spike occurs; otherwise, the accumulated value is retained.

\subsection{Spike-aware Bidirectional Distillation Strategy}
To mitigate temporal mismatch and restore the performance of spiking models under resource constraints, we propose the Spike-aware Bidirectional Distillation Strategy (SBDS). This strategy provides multi-granularity supervision to guide the student model in effectively learning linguistic capabilities from the teacher model. The overall process is illustrated in Figure \ref{fig:distill}.

Specifically, we adopt Llamba as the teacher model, while the student model is a spiking variant constructed using the MAR framework. At the output level, we extend the standard KL loss with a reverse KL divergence term to form a bidirectional distillation loss. While the standard KL emphasizes alignment over the full output distribution, the reverse KL encourages the student model to focus more on high-confidence predictions, making the formulation better suited to the sparse and bursty nature of spiking activations. The corresponding logit alignment loss is defined as follows:
\begin{equation}
    \mathcal{L}_1(p||q) = \sum_{k=0}^{D-1}[\alpha p(k)-\beta q(k)]\cdot[\log p(k)-\log q(k)],
\end{equation}

where $p$ and $q$ denote the teacher and student distributions, respectively, and $D$ is the size of the discrete space, typically the vocabulary size. The parameters $\alpha$ and $\beta$ are tunable hyperparameters that control the trade-off between capturing the overall distribution and emphasizing high-confidence regions, making the loss function better aligned with the sparse characteristics of spiking outputs.

At the feature level, we further design a pre-normalization state alignment loss, which compares the representations of the teacher and student models immediately after the first Root Mean Square Normalization (RMSNorm) module in each layer, thereby facilitating more effective alignment of semantic representations. The formal definition of the loss is given as follows:
\begin{equation}\label{prenorm_loss}
    \mathcal{L}_2(h^j,h^k) = ||\text{PreNorm}(h^j) - \text{PreNorm}(h^s)||_2 ,
\end{equation}
where, $h^j$ and $h^k$ denote the input features at the corresponding layer of the teacher and student models, respectively, and $\text{PreNorm}(\cdot)$ represents the standard pre-layer normalization operation.

Accordingly, the final total loss function is defined as follows:
\begin{equation}\label{total_loss}
    \begin{split}
        \mathcal{L}_{distill} &= \frac{1}{TM}\sum_{t=0}^{T-1}\sum_{m=0}^{M-1}\mathcal{L}_1(p_{t,m}||q_{t,m})\\
        &+\frac{1}{TL}\sum_{t=0}^{T-1}\sum_{l=0}^{L-1}\mathcal{L}_2(h^j_{t,l},h^k_{t,l}),
    \end{split}
\end{equation}
where $T$ is the number of time steps, $M$ is the sequence length, and $L$ is the number of layers. $\mathcal{L}_1$ denotes the logit-level loss between teacher logits $p_{t,m}$ and student logits $q_{t,m}$, while $\mathcal{L}_2$ is the feature-level loss between their pre-normalized hidden states $h_{t,l}^j$ and $h_{t,l}^k$.

This SBDS guides the student model to focus on high-confidence outputs and align semantic representations across layers, accelerating convergence and enhancing performance under resource constraints.

\section{Experiment}
\subsection{Data and experimental settings}

We use Llamba-1B, the Mamba variant of LLaMA-3.2-1B, as the starting point for the second stage of the MAR framework. For training, we employ the GenQA \cite{chen2024genqageneratingmillionsinstructions}, OpenHermes 2.5 \cite{OpenHermes2.5}, and InfinityInstruct \cite{li2025infinityinstructscalinginstruction} datasets, which together contain approximately 7 billion tokens, and train for one epoch.

\subsection{Main results}
\begin{table*}
  \centering
  
  \begin{tabular}{cccllllllll}
    \hline
    \multirow{2}{*}{Model} & \multirow{2}{*}{Size} & \multirow{2}{*}{SNN} & \multicolumn{7}{c}{Accuracy (\%)} \\ \cline{4-10}
    & & & PIQA & BoolQ & WG & HS & ARC-e & ARC-c & Average \\ \hline
    LLaMA & 1.3B & no  & 74.32 & 69.54 & 59.67 & 60.76 & 68.52 & 38.05 &  61.80\\
    Llamba & 1.4B & no  & 73.78 & 68.62 & 60.69 & 61.31 & 69.57 & 37.32 & 61.88 \\
    Bi-Mamba\cite{tang2024bi} & 1.3B & no & 69.20 & 62.00 & 53.70 & 43.10 & 43.90 & 24.40 & 49.38 \\
    SmoothQuant\cite{xiao2023smoothquant} & 1.3B & no  & 71.60 & 57.77 & 60.06 & 53.68 & 57.03 & 29.44 & 54.93 \\
    TinyLLaMA\cite{zhang2024tinyllama} & 1.3B & no & 73.30 & 57.80 & 59.10 & 59.90 & 55.30 & 30.10 & 55.91 \\ 
    SpikeLLM\cite{xing2024spikellm} & 7B & yes  & 65.45 & 64.37 & 54.3 & 56.59 & 41.67 & 32.51 & 52.48 \\
     Ours & 1.4B & yes  & 70.35 & 68.17 & 55.33 & 50.91 & 67.21 & 35.58 & 57.20 \\ \hline
\end{tabular}
\caption{Comparison of zero-shot performance on commonsense reasoning and question answering benchmarks across various models. For ARC-E and HS, we use normalized logits’ results.}
\vspace{-5pt}
  \label{tab:commenSense}
\end{table*}
To comprehensively assess both the effectiveness and efficiency of our proposed MAR framework, we evaluate the model on two fronts: accuracy performance across zero-shot reasoning benchmarks, and energy consumption during inference.

We conduct zero-shot evaluations on six commonsense reasoning and question answering datasets: Physical Interaction QA (PIQA) \cite{bisk2020piqa}, Boolean Questions (BoolQ) \cite{clark2019boolq}, Winogrande (WG) \cite{sakaguchi2020winogrande}, HellaSwag (HS) \cite{zellers2019hellaswag}, ARC Challenge (ARC-C), and ARC Easy (ARC-E)\cite{clark2018thinksolvedquestionanswering}. As shown in Table \ref{tab:commenSense}, our method achieves an average accuracy of 57.93\% across six tasks. Compared with the teacher model Llamba (61.88\%), its performance is largely restored; compared with the much larger SpikeLLM (7B, 52.48\%), our model attains a 5.45 percentage point improvement using only 1.4B parameters, demonstrating superior parameter efficiency. In addition, it also surpasses other efficiency-oriented models, such as Bi-Mamba (49.38\%), SmoothQuant (54.93\%), and TinyLLaMA (55.91\%), confirming that the proposed approach can effectively recover performance under limited resource conditions.

\begin{figure}[ht]
    \centering
    \includegraphics[width=0.9\columnwidth]{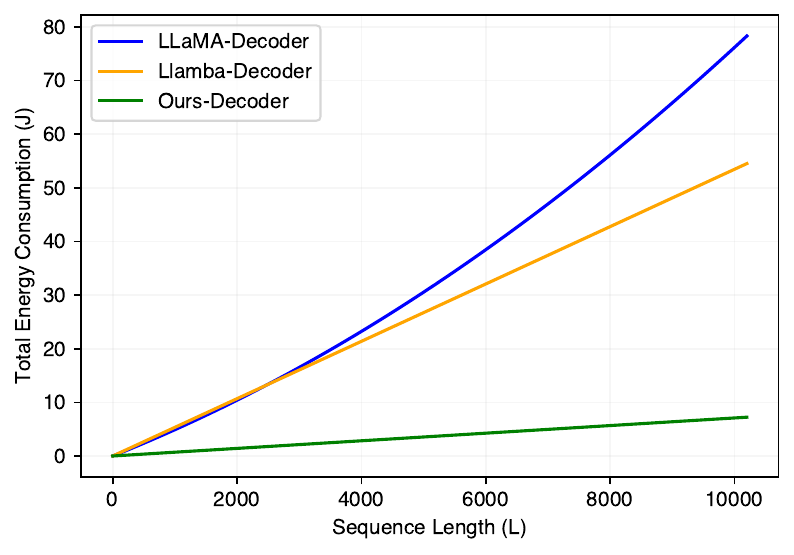}
    \vspace{-5pt}
    \caption{Total energy consumption of different decoders at varying sequence lengths.}
    \vspace{-5pt}
    \label{fig:energyComparison}
\end{figure}

To further quantify the efficiency advantages of the proposed method, we compare the inference energy consumption of the MAR model with its dense baseline. Following the energy metrics reported by \cite{horowitz20141}, we assume that each MAC, multiplication, and AC operation consumes 4.6, 3.7, and 0.9 pJ, respectively. On the WG dataset, we measure the average firing rate of spiking neurons in the decoder and estimate the total energy consumption at different sequence lengths. As shown in Figure \ref{fig:energyComparison}, compared with same-sized Llamba and LLaMA models, MAR exhibits a substantially slower growth in energy consumption. Moreover, MAR consistently incurs lower inference energy across all sequence lengths, and this advantage becomes more pronounced as the sequence length increases. These results demonstrate the superiority of MAR in effectively reducing the computational energy cost of FFNs and other linear layers.

\subsection{Ablation study}
To assess the individual contributions of each component within the MAR framework, we conduct a series of systematic ablation studies. Each subsection provides a focused analysis of how different modules influence overall model performance, with the average accuracy across six benchmark tasks used as the evaluation metric.

As shown in Table \ref{tab:ablation}, we progressively introduce ATMN, the reverse KL divergence compensatory term, and the pre-normalization alignment loss on a shared baseline to evaluate the contribution of each module to model performance. The baseline, which contains only traditional binary neurons with standard KL-based distillation, achieves an average accuracy of 46.28\%. Replacing the binary neurons with ATMN increases accuracy to 55.20\%, demonstrating its effectiveness in alleviating the issue of reduced information density. Adding the reverse KL term further improves accuracy to 55.46\%, indicating that this compensatory component helps address the sparse and bursty nature of spiking outputs. With the additional introduction of pre-normalization alignment loss, accuracy rises to 57.20\%, highlighting enhanced representational consistency between teacher and student models. These results collectively validate the effectiveness of the proposed optimization strategy.

\begin{table}[h]
  \centering
  
  \begin{tabular}{cccc}
  \hline
    ATMN       &Reserve KL & PN loss             & Average acc. (\%) \\ \hline
    -          &-          &-                    & 46.28             \\
    \checkmark &-          &-                    & 55.20             \\
    \checkmark &\checkmark & -                   & 55.46             \\
    \checkmark &\checkmark & \checkmark          & 57.20             \\ \hline
  \end{tabular}
  \caption{Ablation study of core components. Each column shows whether a module is enabled. Reserve KL and PN loss denote the reverse KL divergence term and the pre-normalization alignment loss.}
  \vspace{-5pt}
  \label{tab:ablation}
\end{table}

Next, we compare three feature-level alignment strategies to examine their role in the distillation process: using only pre-normalization alignment, using only post-normalization alignment, and combining both. The baseline configuration includes only logit-level distillation loss, and all strategies are evaluated on this basis. As shown in Table \ref{tab:PreNormAblation}, pre-normalization alignment achieves the best performance, while post-normalization alignment alone or the combination of both yields relatively lower accuracy. This indicates that pre-normalization features provide more stable and transferable representations for guiding the student model. Therefore, we adopt pre-normalization alignment as the default feature-level distillation strategy in our final framework.
\begin{table}[h]
  \centering
  \begin{tabular}{ccc}
    \hline
    Pre-norm   & Post-norm  & Average acc. (\%) \\ \hline
    -          & -          & 55.46             \\
    \checkmark & -          & 57.20             \\
    -          & \checkmark & 56.75             \\
    \checkmark & \checkmark & 56.08             \\ \hline
  \end{tabular}
  \caption{Comparison of feature-level alignment strategies. Pre-Norm aligns features after the first RMSNorm in each layer, whereas Post-Norm aligns features after the final normalization.}
  \vspace{-5pt}
  \label{tab:PreNormAblation}
\end{table}

In addition, we conducted a sensitivity analysis of the hyperparameters $\alpha$ and $\beta$ to examine the effect of weighting configurations in SBDS. As shown in Table \ref{tab:ab-ablation}, using only standard KL ($\alpha=1,\ \beta=0$, 55.20\%) or only reverse KL ($\alpha=0,\ \beta=1$, 55.92\%) failed to achieve optimal performance. By contrast, combining the two led to significant improvements, with the best result obtained at $\alpha=0.2,\ \beta=0.7$, reaching an average accuracy of 56.40\%. These results confirm the effectiveness of the bidirectional distillation loss.
\begin{table}[h]
  \centering
  
  \begin{tabular}{ccc}
    \hline
    $\alpha$ & $\beta$ & Average acc.(\%) \\ \hline
    0       & 1      & 55.92        \\
    1       & 0      & 55.20            \\
    0.7     & 0.2    & 56.39        \\
    0.2     & 0.7    & 56.40        \\ \hline
  \end{tabular}
  \caption{Sensitivity analysis of hyperparameters $\alpha$ and $\beta$ in SBDS.}
  \vspace{-5pt}
  \label{tab:ab-ablation}
\end{table}

\section{Conclusion}
This paper proposes a two-stage MAR framework to jointly optimize attention mechanisms and FFNs in LLMs. In the first stage, SSMs are introduced to achieve linear-time sequence modeling; in the second, activation spiking is applied to reduce the computational cost of FFNs. In addition, ATMN is designed to mitigate the issue of low information density, while SBDS addresses temporal mismatches and restores performance. Experimental results show that the proposed method effectively recovers teacher-level performance under constrained resources, substantially reduces inference energy consumption, and outperforms not only models of comparable scale but also larger efficient models, demonstrating its potential to achieve a favorable balance between accuracy and efficiency.



\bibliographystyle{IEEEbib}
\bibliography{strings,refs}

\end{document}